\documentclass[runningheads]{llncs}
\usepackage{graphicx}
\usepackage{amsfonts}
\usepackage{multirow}
\usepackage{amsmath}
\usepackage[table,xcdraw]{xcolor}
\usepackage{tabularray}
\usepackage{threeparttable}
\usepackage{booktabs}
\usepackage{multirow}
\usepackage{dsfont}
\usepackage{tabularx}
\usepackage{hyperref}

\usepackage{xr-hyper}
\makeatletter

\newcommand*{\addFileDependency}[1]{
\typeout{(#1)}
\@addtofilelist{#1}
\IfFileExists{#1}{}{\typeout{No file #1.}}
}\makeatother

\newcommand*{\myexternaldocument}[1]{%
\externaldocument{#1}%
\addFileDependency{#1.tex}%
\addFileDependency{#1.aux}%
}
\myexternaldocument{supplementary}

\begin{document}

\title{HuLP: Human-in-the-Loop for Prognosis}

\titlerunning{Human-in-the-Loop for Prognosis}

\author{Muhammad Ridzuan \and Mai A. Shaaban \and Numan Saeed \and Ikboljon Sobirov \and Mohammad Yaqub} 

\authorrunning{M. Ridzuan et al.}

\institute{Mohamed Bin Zayed University of Artificial Intelligence, Abu Dhabi, UAE \\
\email{\{Muhammad.Ridzuan, Mai.Kassem, Numan.Saeed, Ikboljon.Sobirov, Mohammad.Yaqub\}@mbzuai.ac.ae}}

\maketitle              

\begin{abstract}
This paper introduces HuLP, a Human-in-the-Loop for Prognosis model designed to enhance the reliability and interpretability of prognostic models in clinical contexts, especially when faced with the complexities of missing covariates and outcomes. HuLP offers an innovative approach that enables human expert intervention, empowering clinicians to interact with and correct models' predictions, thus fostering collaboration between humans and AI models to produce more accurate prognosis. Additionally, HuLP addresses the challenges of missing data by utilizing neural networks and providing a tailored methodology that effectively handles missing data. Traditional methods often struggle to capture the nuanced variations within patient populations, leading to compromised prognostic predictions. HuLP imputes missing covariates based on imaging features, aligning more closely with clinician workflows and enhancing reliability. We conduct our experiments on two real-world, publicly available medical datasets to demonstrate the superiority and competitiveness of HuLP. Our code is available at \url{https://github.com/BioMedIA-MBZUAI/HuLP}.

\end{abstract}

\section{Introduction}
Diagnosis and prognosis play pivotal roles in oncology, yet prognosis presents a unique challenge due to its heightened uncertainty and complex nature. Unlike diagnosis, which primarily focuses on confirming the presence of cancerous cells or tumors \cite{thakor2013nanooncology}, prognosis entails predicting the trajectory of the disease, including survival time and likelihood of recurrence \cite{glare2008predicting}. This complexity arises from various factors that influence disease progression and outcome, ranging from tumor characteristics to patient demographics and treatment efficacy \cite{zugazagoitia2016current}, making prognosis more challenging for clinicians to assess accurately. 

While deep learning models are emerging as clinical assistants in prognosis, current approaches face two significant problems in the clinical setting. First, the models leave no space for clinicians to intervene, even when the models are incorrect or less confident, thus limiting the clinicians' ability to provide valuable inputs or improve the models' predictions. Related works allowing human intervention \cite{cem_hybrid_icml2021,cem2022} are applied to natural images but not used for prognosis. During inference, the models can benefit from such feedback to improve their overall performance, mimicking how doctors collaborate and refine their assessment based on collective expertise. Presently, there is a gap in established methodologies (e.g., \cite{deephit,deepsurv,ensemble_late_fusion}) that enable active human interaction and intervention to refine the model's predictions of clinical features to improve prognosis.

Second, in cancer prognosis, dealing with incomplete data and censored patient outcomes (i.e., instances for which we do not know the exact event time) is challenging. Missing covariates may result from incomplete collection  \cite{janssen2010missing}, non-compliance \cite{levy2004covariate}, or technical errors \cite{janssen2010missing}, while missing outcomes may arise due to patients discontinuing follow-up visits \cite{ensemble_late_fusion}, relocating \cite{sparr1993returns}, or withdrawing from a study \cite{sparr1993returns}. Standard practice in AI research typically uses naive imputation methods such as statistical measures of central tendency (i.e., mean, median, mode), $k$-nearest neighbor, or more algorithmic approaches, such as multiple imputation by chained equations (MICE). However, in reality, oncologists rely on radiological images to gain more insights into the patients' conditions  \cite{hosny2018artificial}.

The use of electronic health records (EHR) alone in prognosis often falls short of capturing the complex variability among individuals within and across different medical contexts, especially in static non-temporal EHR datasets. For instance, consider two individuals with identical clinical profiles, both diagnosed with lung cancer; despite sharing similar clinical information, their survival outcomes can exhibit significant disparities. Table \ref{table:ehr_variability} highlights several such real-world cases from the ChAImeleon \cite{CHAIMELEON} lung cancer dataset. Traditional models trained solely on EHR data struggle to reliably distinguish such variations in survival. Notably, the integration of radiological images -- which provides a richer manifestation of temporal information, including age, smoking status, and tumor texture and characteristics -- offers a promising avenue for capturing data dynamics that are often overlooked in static clinical data.

In response to these challenges, we introduce \textbf{Human-in-the-Loop for Prognosis (HuLP)}, a deep learning architecture inspired by \cite{cem_hybrid_icml2021,cem2022} designed to enhance the reliability and interpretability of prognostic models in clinical settings. Our main contributions are twofold:
\begin{itemize}
         \item \textit{Allowing user interaction and expert intervention at inference time:} HuLP facilitates human expert intervention during model inference, empowering clinicians to provide input and guidance based on their domain expertise. This capability significantly enhances the model's decision-making process, particularly in complex prognostic scenarios where expert knowledge is invaluable.
         \item \textit{Capability of handling both missing covariates and outcomes and extraction of meaningful vector representations for prognosis:} HuLP is equipped with a robust mechanism for handling missing data, ensuring end-to-end reliability in prognostic predictions. By leveraging patients' clinical information as intermediate concept labels, our model generates richer representations of clinical features, thereby enhancing prognostic accuracy.
\end{itemize}

\begin{table}[!t] 
\centering
\caption{Variability in patient survival times from the ChAImeleon~\cite{CHAIMELEON} lung cancer dataset for a given set of identical covariates. Event “1" signifies the patient's death at the given time (\textit{uncensored}); event “0" signifies that the patient is alive at (least until) the given time (\textit{censored}). “X" represents unknown or missing data.} 
\label{table:ehr_variability}
\resizebox{.95\textwidth}{!}{%
    \begin{tabular}{cccccccc}
        \toprule
        \textbf{Age} $|$ & \textbf{Gender} $|$ & \textbf{Smoking Status} $|$ & \textbf{T-stage} $|$ & \textbf{N-stage} $|$ & \textbf{M-stage} $|$ & \textbf{Survival (months)} $|$ & \textbf{Event} \\
        \midrule
        \multicolumn{8}{l}{Patients with the same TNM} \\
        \midrule
        70 & Male & Ex-smoker & T4 & N3 & M1c & 3.50 & 1 \\
        70 & Male & Ex-smoker & T4 & N3 & M1c & 1.20 & 1 \\
        \midrule
        \multicolumn{8}{l}{Patients with missing TN} \\
        \midrule
        72 & Male & Ex-smoker & X & X & M1 & 15.23 & 0 \\
        72 & Male & Ex-smoker & X & X & M1 & 5.17 & 0 \\
        \midrule
        \multicolumn{8}{l}{Patients with missing TNM} \\
        \midrule
        57 & Male & Smoker & X & X & X & 3.50 & 1 \\
        57 & Male & Smoker & X & X & X & 56.53 & 0 \\
        67 & Female & Ex-smoker & X & X & X & 4.27 & 0 \\
        67 & Female & Ex-smoker & X & X & X & 58.27 & 0 \\

        \bottomrule
    \end{tabular}%
}
\end{table}

\begin{figure}[!t]
    \centering
    \includegraphics[scale=0.13]{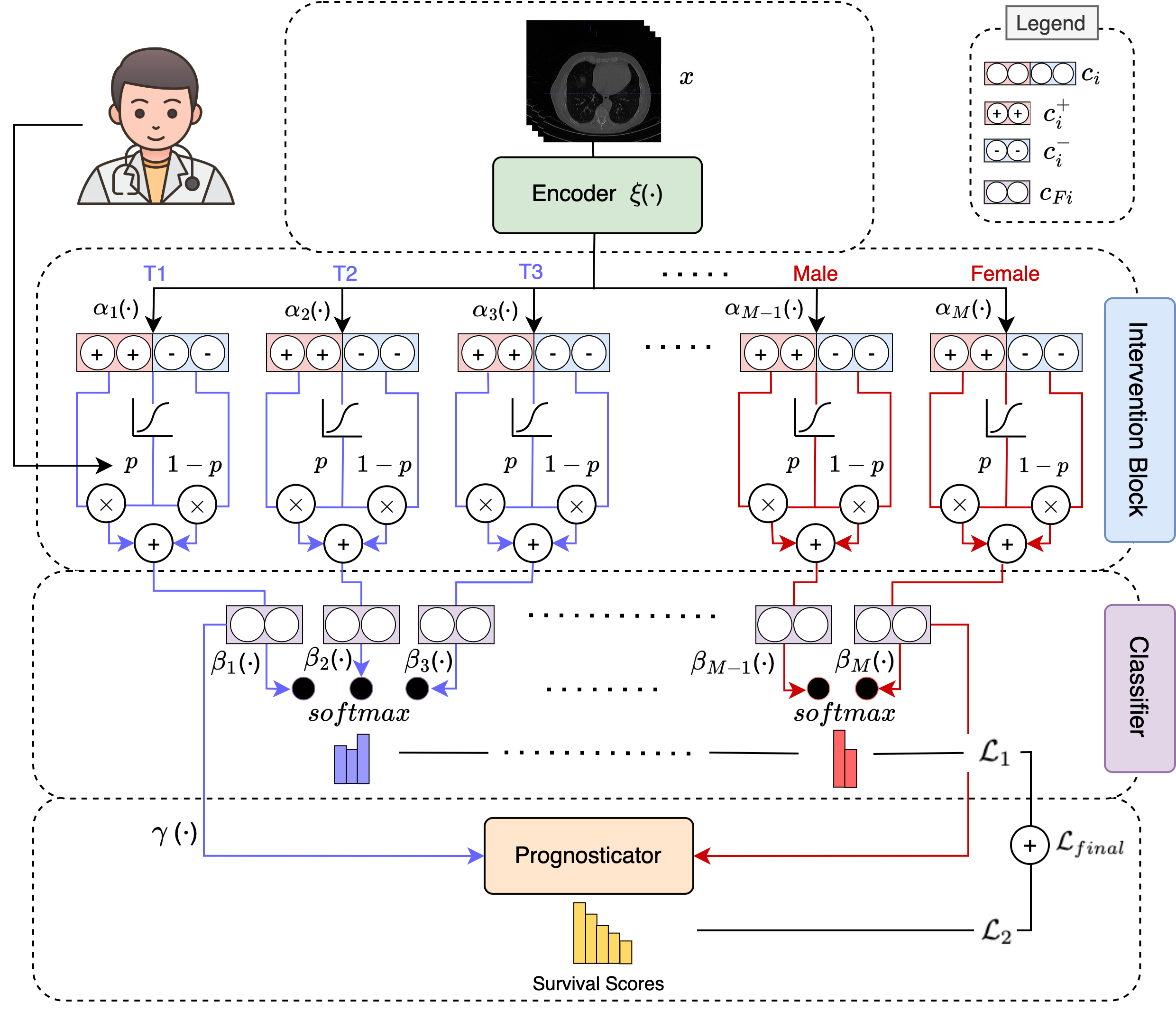}
    \caption{HuLP is composed of (1) a deep learning \textbf{encoder} that extracts features from medical images; (2) an \textbf{intervention block} that allows human intervention during test time; (3) a \textbf{classifier} that ensures concept alignment of the feature embeddings; and (4) a \textbf{prognosticator} that performs survival prediction. Here, \textit{(T1, T2, T3)} and \textit{(Male, Female)} are example clinical concepts obtained from the parent categories \textit{T-stage} and \textit{gender}, respectively. Our loss is a combination of the concept loss \textbf{$L_1$} applied on the classifier and prognosis loss \textbf{$L_2$} applied on the prognosticator.} 
    \label{Figure:model}
\end{figure}

\section{Methodology}

Figure~\ref{Figure:model} shows the complete architecture of the proposed HuLP model, which is comprised of four main components: \textbf{encoder}, \textbf{intervention block}, \textbf{classifier}, and \textbf{prognosticator}. 

The \textbf{encoder} $\xi(\cdot)$ is a deep neural network (e.g. CNN or transformer) that processes an image $x\in\mathbb{R}^{H\times W\times D\times C}$, where $H, W, D$ and $C$ are the height, width, depth, and number of channels of the input, respectively, and generates a latent feature embedding $y\in\mathbb{R}^{(K \times M)}$. Here, $K$ represents the embedding space dimension, while $M$ is the total number of unique and discrete patient characteristics (\textit{concepts}) across all clinical features (\textit{parent categories}) $P$ in EHR, i.e. $M = \sum_{j=1}^{|P|} m_j$, where $m_j$ denotes the number of unique concepts $m$ in each feature $j$. Continuous features are discretized. This embedding output plays a foundational role in capturing essential features from the image and is designed to learn a shared representation. It is then passed through $M$ groups of fully connected (FC) layers $\alpha(\cdot)$ to produce concept embeddings $c$.

The \textbf{intervention block}, a key component of HuLP, enables user interaction and expert intervention at test time. During training, each group of embeddings $c_i=\alpha_i(y)$ undergoes a single-neuron sigmoid weighting function, producing the probability $p$ of a concept being active ($p_i=\sigma(c_i)$). The embeddings are then split arbitrarily into two halves, $c_i^+$ and $c_i^-$, and multiplied by $p_i$ and $(1-p_i)$ to represent the latent positive and negative concept embeddings, respectively. To facilitate test-time intervention, $p$ is randomly replaced with the hard ground-truth labels $[0,1]$ with a probability of 0.25. During inference, a human may impart his/her domain knowledge by replacing $p$ with [0,1] to indicate certainty in the presence or absence of a concept. The final concept embedding is generated as the sum of the positive and negative embeddings ($c_{Fi} = p_i c_i^+ + (1-p_i) c_i^-$). 

The \textbf{classifier} is an FC layer $\beta (\cdot)$ that encourages concept alignment of the embeddings by enforcing each embedding layer to predict only one concept (i.e. $\beta_i(c_{Fi}) \in\mathbb{R}^1$). 
This is followed by a softmax and cross-entropy loss for the concepts under each parent category. The classifier is designed to allow missing data (see \textit{Loss function} for details). 

Finally, the \textbf{prognosticator} processes the concepts $w=[c_{F1}, c_{F2}, ..., c_{FM}]$ through an FC layer $\gamma (\cdot)$ with $n$ discrete time bins. The softmax outputs of the layer represent the estimated hazard for each patient, indicating the instantaneous rate of an event (e.g., death or cancer recurrence) conditioned on surviving up to time $t$ for a given patient concept vector $w$. The prognosticator layer is responsible for predicting the progression of clinical outcomes over time. The pseudo-code of the model is described in Algorithm 1 in the \textit{Appendix}.

\textbf{Loss function.}
\label{loss}
The proposed loss is a combination of the concept loss $\mathcal{L}_1$ and prognosis loss $\mathcal{L}_2$. The concept loss $\mathcal{L}_1$ is applied on the classifier layer using cross-entropy and is computed as an average over all non-missing covariates (Eq. \ref{eq:XE}). This loss is backpropagated for patients with non-missing covariates but skipped for patients with missing covariates, thus providing the advantage of avoiding hard imputation of missing data prior to training.

\begin{equation}
\label{eq:XE}
\mathcal{L}_1 = -\frac{1}{\hat{N}} \sum_{i=1}^{M} \sum_{j=1}^{\hat{N}} y_{ij} \log(\beta_i(c_{Fij}))
\end{equation}

where $M$ is the number of discrete clinical categories (see \textit{encoder}) and $\hat{N}$ is the number of patients with non-missing data. 

The prognosis loss $\mathcal{L}_2$ is applied to the prognosticator. It is a slightly modified version of the DeepHit \cite{deephit} loss function for a single non-competing risk. Given time $T$, event indicator $E$, and concept vector $w$, we convert the outputs of HuLP into an estimated survival function $\hat{S}$ using

\begin{equation}
\label{eq:survival_cum_hazard}
\hat{S}(T \mid w) = exp(-\hat{H}(T \mid w)) = exp(-\sum_{t=1}^{T} \hat{h}(t \mid w))
\end{equation}

where $\hat{H}$ is the cumulative hazard function and $\hat{h}$ is the estimated hazard from the softmax outputs of the model. $\mathcal{L}_2$ is thus defined as a weighted average of the discrete log-likelihood and rank losses: 

\begin{equation}
\label{eq:L2}
\mathcal{L}_2 = a \text{loss}_{\mathcal{LL}} + (1 - a) \text{loss}_{\text{rank}}\\
\end{equation}

where 
\begin{equation}
\label{eq:loglikelihood_loss}
{loss}_{\mathcal{LL}} = - \sum_{i=1}^{N} \left[ E_i \log(\hat{h}_{e_i}(T_i \mid w_i) + (1 - E_i) \log(\hat{S}(T_i \mid w_i)) \right]\\
\end{equation}

and
\begin{equation}
\label{eq:rank_loss}
{loss}_{\text{rank}} = \sum_{i,j} E_i \, \mathds{1}\{T_i < T_j\} \exp\left(\frac{\hat{S}(T_i \mid w_i) - \hat{S}(T_j \mid w_j)}{c}\right)\\
\end{equation}

$N$ is the total number of patients, $e_i$ is the index of the event time for observation, and $c$ is set to a constant 0.1 following \cite{deephit}.

The log-likelihood (Eq. \ref{eq:loglikelihood_loss}) captures information regarding the time of the event and its occurrence for uncensored patients, and the time at which the patient was lost to follow-up (indicating that the patient was alive up to that time) for censored patients. The ranking loss (Eq. \ref{eq:rank_loss}) compares the survival scores between possible pairs $i, j$ of patients to incentivize the correct ordering of pairs. 

The final loss is calculated using:

\begin{equation}
\label{eq:final}
\mathcal{L}_{final} = b \mathcal{L}_1 + (1 - b) \mathcal{L}_2\\
\end{equation}

$a$ (in Eq. \ref{eq:L2}) and $b$ are weighting hyperparameters.

\section{Experimental Setup}
\subsection{Datasets}
The prognostic ability of HuLP is assessed by comparing it with conventional benchmarks in analyzing two real-world medical datasets: ChAImeleon \cite{CHAIMELEON} and HECKTOR \cite{hecktor2021a,hecktor2021b}. Below, we provide a brief overview of each.

The ChAImeleon \cite{CHAIMELEON} lung cancer dataset consists of 320 patient CT scans with EHR. The clinical features include age, gender, smoking status, clinical category (T-stage), regional nodes category (N-stage), and metastasis category (M-stage), with up to 26\% missing and 59\% censored data. For preprocessing, we combined all missing data labels, i.e. “Unknown”, nan, “TX”, and “NX” into the same category “X” for each feature. All cancer sub-stages are combined into their parent stage to increase the number of samples per category
(e.g. T1a, T1b, T1c are combined into T1). We use a publicly available segmentation model \cite{Hofmanninger2020} to restrict the ROI to the lung areas. 

HECKTOR \cite{hecktor2021a,hecktor2021b} is a multi-modal head-and-neck cancer dataset comprising 224 CT and PET scans with EHR. The PET and CT scans are registered to a common origin. The clinical features include center, age, gender, TNM 7/8th edition staging and clinical stage, tobacco and alcohol consumption, performance status, HPV status, and treatment (chemoradiotherapy or radiotherapy only), with up to 90\% missing and 75\% censored data. Features with over 80\% missing data are dropped, and all cancer sub-stages are combined into their parent stage. 
To standardize the inputs, the scans are preprocessed in the same manner for each dataset via resampling, cropping, and resizing.

\subsection{Implementation Details}
HuLP is run for 100 epochs using DenseNet-121 \cite{densenet} as the encoder. We use positive/negative embeddings of size 64, combined to form a final concept embedding of size 32. The prognosticator outputs 12 discrete time bins for ChAImeleon \cite{CHAIMELEON} and 16 for HECKTOR \cite{hecktor2021a,hecktor2021b}, obtained as the square root of the number of observations corresponding to the quantiles of the survival time distribution. We use a batch size of 32, AdamW \cite{adamw} optimizer with a learning rate of $1\times10^{-3}$, and a cosine annealing scheduler with a warmup of 5 epochs. All experiments are implemented using PyTorch \cite{pytorch}. 

We compare HuLP against three deep survival methods: DeepHit \cite{deephit}, Deep-MTLR \cite{mtlr}, and Fusion \cite{ensemble_late_fusion}. DeepHit \cite{deephit} and Deep-MTLR \cite{mtlr} are chosen because they are both top-performing discrete survival methods in prognosis; similarly, our HuLP implementation is also discrete. Fusion \cite{ensemble_late_fusion} is chosen as a multimodal baseline and also because it won the HECKTOR \cite{hecktor2021a,hecktor2021b} competition, the same dataset used in this work. 

DeepHit \cite{deephit} and Deep-MTLR \cite{mtlr} (EHR) are run using mode imputation with two FCs of size 64 each followed by a ReLU activation, batch normalization and dropout with probability 0.1, and a prognosticator using a batch size of 96 and learning rate of $1\times10^{-2}$. We also compare our method against a variant of DeepHit \cite{deephit} and DeepMTLR \cite{mtlr} using imaging data as inputs and DenseNet-121 \cite{densenet} as the encoder to directly predict survival outcomes. Finally, we implement the idea of Fusion \cite{ensemble_late_fusion} by extracting imaging features using DenseNet-121 \cite{densenet}, concatenated with EHR through a late fusion technique. We maintain a constant ratio of patients who experienced each event and those who were censored in each fold. The experiments are repeated with two seeds and five-fold cross-validation.

\section{Results}
We report the time-dependent concordance (C-index) of Antolini et al. \cite{antolini} from the survival curves. Table \ref{table:main2} summarizes our results. EHR without images presents the problem of limited depth and richness of static information. Images without EHR leave the model unguided. HuLP consistently demonstrates statistically significant improvements ($p$-value\textless 0.05) over these methods and remains competitive with Fusion \cite{ensemble_late_fusion}; however, 
the learning of fusion from EHR and image embeddings was disjoint. HuLP distinguishes itself by integrating EHR as an intermediate concept labeling that guides the model towards the relevant features, thus producing rich, disentangled embeddings of the clinical features from the images with two added advantages: it allows human expert intervention during test time and is robust to missing data.

To emulate human intervention, $p$ is fully replaced with ground-truth labels for non-missing data, while $p$ is retained for missing data. We run inference on the validation set with and without test-time intervention. 
Notably, the integration of user interaction and expert intervention of the clinical concepts significantly enhances the model's prognostic capabilities (Table \ref{table:testtime}), yielding an improvement of about 0.1 C-index on ChAImeleon \cite{CHAIMELEON}. 

\begin{table}[!b] 
\centering
\caption{Average concordance indices on two seeds and five-fold cross-validation. The highest scores per dataset are bolded. (*) is shown for statistically significant experiments ($p$-value $<0.05$) based on the average performance of HuLP and the best-performing baseline. 
} \label{table:main2}
    \begin{tabularx}{0.95\columnwidth}{@{\extracolsep{\fill}} lccc }
        \toprule
        & Modality & ChAImeleon Lung Cancer~\cite{CHAIMELEON} & HECKTOR~\cite{hecktor2021a,hecktor2021b} \\
        \midrule
        \textbf{DeepHit~\cite{deephit}} & EHR & 0.6522* $\pm$ 0.0371 & 0.6054* $\pm$ 0.1047 \\
        \textbf{DeepMTLR~\cite{mtlr}} & EHR & 0.6624* $\pm$ 0.0643 & 0.6085* $\pm$ 0.0985  \\
        \midrule
        \textbf{DeepHit~\cite{deephit}} & Image & 0.6328* $\pm$ 0.0559 & 0.7144 $\pm$ 0.0269 \\
        \textbf{DeepMTLR~\cite{mtlr}} & Image & 0.6400* $\pm$ 0.0361 & 0.6222* $\pm$ 0.0788 \\
        \midrule
        \textbf{Fusion~\cite{ensemble_late_fusion}} & Image+EHR & \textbf{0.7399} $\pm$ 0.0534 & 0.7012 $\pm$ 0.0457 \\
        \textbf{HuLP (ours)} & Image+EHR & 0.7124 $\pm$ 0.0533 & \textbf{0.7329} $\pm$ 0.0415 \\ 
        \bottomrule
    \end{tabularx}
\end{table}

\begin{table}[t!] 
\centering
\caption{Effect of test-time concept interventions on concordance index scores for ChAImeleon \cite{CHAIMELEON}. The results shown are for five-fold cross-validation with two seeds.} \label{table:testtime}
    \begin{tabularx}{0.6\columnwidth}{@{\extracolsep{\fill}} cc }
        \toprule
        With test-time interv. & Without test-time interv. \\
        \midrule
        0.7124 $\pm$ 0.0533  & 0.6060 $\pm$ 0.0441 \\

        \bottomrule
    \end{tabularx}
\end{table}

\begin{table}[t!] 
    \centering
    \caption{Effect of different imputation methods on concordance index scores on ChAImeleon \cite{CHAIMELEON}. The results shown are the averages of three seeds. The highest scores per column are bolded.}  
    \label{table:impute}
    \begin{tabularx}{0.8\columnwidth}{@{\extracolsep{\fill}} l|cccc }
        \toprule
        \multicolumn{5}{c}{Missing data percentage} \\
        \midrule
        Imputation & 30\% & 40\% & 50\% & 70\% \\
        \midrule
        Mode & 0.5817 & 0.6563 & 0.6401 & 0.6430 \\
        kNN (k=1) & 0.6068 & 0.6526 & 0.6556 & 0.6585 \\
        MICE & 0.6068 & 0.6275 & 0.6541 & 0.6371 \\
        HuLP (ours) & \textbf{0.6297} & \textbf{0.6748} & \textbf{0.6740} &  \textbf{0.6593} \\
        
        \bottomrule
    \end{tabularx}
\end{table}

To investigate HuLP's robustness to missing data, 
we create a challenging 8:2 train-validation split stratified by gender where each patient in the validation split has identical or similar EHR as at least one other patient and at least one missing data. We randomly mask entries in the training EHR with increasing probabilities to emulate situations with missing data on the ChAImeleon \cite{CHAIMELEON} dataset. Table A1 in the \textit{Appendix} details the distribution of the validation split. 
We run our experiments for three seeds and compare our method against conventional benchmarks for imputation, including mode, kNN, and MICE. Table \ref{table:impute} presents the results of our experiments, showing our method's robustness to missing data, particularly in the low missing-data regime. At high missing-data regime, the improvement becomes less significant, likely because the model receives inadequate feedback from $\mathcal{L}_1$ to capture the semantic meaning of the concepts. However, the results remain competitive with the baselines.

\section{Discussion}
To our knowledge, HuLP is the first prognostic model that allows human interaction and intervention of known concepts for prognosis. This innovation represents a significant advancement particularly in prognosis, where predicting future outcomes can often be more challenging than diagnosing present conditions. Compared to methods where human experts are passive users, HuLP empowers clinicians to actively engage with the model, refining its concept predictions based on their domain expertise. This collaborative approach fosters a synergistic relationship between humans and computers, allowing each to leverage their strengths. Clinicians, with their deep understanding of clinical features, can provide refined adjustments to the model's predictions regarding the presence or absence of certain concepts, while HuLP dynamically incorporates these inputs to enhance the accuracy of prognostic assessments. This active collaboration not only improves the interpretability and reliability of prognostic models but also instills confidence in their use in clinical decision-making.

Additionally, in addressing the challenge of missing data, HuLP presents a tailored methodology that surpasses conventional approaches like mode, kNN, and MICE. These methods, while widely used, often oversimplify the complexity of clinical datasets and may introduce bias, thereby compromising the validity of prognostic predictions. In contrast, HuLP harnesses the power of neural networks to better accommodate the nuances of missing data in prognostic modeling. In particular, during test time, HuLP implicitly imputes the missing covariates based on the imaging features rather than relying on a simplistic hard imputation. This aligns more closely with clinician workflows and enhances the reliability and trustworthiness of prognostic assessments.

\section{Conclusion}
This paper presents HuLP, Human-in-the-Loop for Prognosis, an innovative approach that allows clinicians to interact with and intervene in model predictions at test time, enhancing prognostic model reliability and interpretability in clinical settings. HuLP extracts meaningful representations from imaging data and can effectively handle missing covariates and outcomes. Experimental results on two medical datasets demonstrate HuLP's superior and competitive performance. Future work should focus on validating HuLP in clinical settings with clinical inputs and exploring the usability of the disentangled feature embeddings.

\begin{credits}
\subsubsection{\ackname} De-identified data used in the development of this solution were prepared and provided by the ChAImeleon Project within the Open Challenges organized by its Consortium, funded by the European Union’s Horizon 2020 research and innovation programme under grant agreement No. 952172, entitled ‘Accelerating the lab to market transition of AI tools for cancer management.

\end{credits}

\bibliographystyle{splncs04}
\bibliography{main.bib}

\end{document}


\begin{appendix}
\section{HULP: Human-in-the-Loop for Prognosis (Appendix)}
\begin{table}[] 
\centering
\caption{Distribution of custom validation split (64 patients) from the ChAImeleon \cite{CHAIMELEON} lung cancer dataset for the imputation experiments. Survival range is the recorded time in months. “X" represents unknown or missing data. Count is the count of patients with the same clinical information.} 
\label{table:val_impute_split}
\resizebox{1.0\textwidth}{!}{%
    \begin{tabular}{ccccccc}
        \toprule
        \textbf{Count} $|$ & \textbf{Gender} $|$ & \textbf{Smoking Status} $|$ & \textbf{T-stage} $|$ & \textbf{N-stage} $|$ & \textbf{M-stage} $|$ & \textbf{Surv. Range (months)} \\
        \midrule
        10 & Female & Smoker & X & X & X & 4.43 to 52.37 \\
        5 & Female & Smoker & X & X & cM0 & 8.60 to 61.00 \\
        4 & Female & Ex-smoker & X & X & X & 4.27 to 58.27 \\
        3 & Female & X & cT4 & cN2 & cM1 & 7.40 to 28.97 \\
        3 & Female & Non-smoker & X & X & X & 2.33 to 24.50 \\
        2 & Female & X & cT3 & cN2 & cM1 & 6.50 to 28.60 \\
        2 & Female & Ex-smoker & X & X & cM1 & 10.40 to 26.47 \\
        1 & Female & Smoker & X & X & cM1 & 0.83 to 0.83 \\
        1 & Female & Ex-smoker & X & X & cM0 & 36.17 to 36.17 \\
        \midrule
        9 & Male & Ex-smoker & X & X & X & 2.27 to 57.10 \\
        7 & Male & Smoker & X & X & X & 2.57 to 56.53 \\
        6 & Male & Non-smoker & X & X & X & 2.83 to 23.80 \\
        4 & Male & Smoker & X & X & cM1 & 5.63 to 20.27 \\
        4 & Male & Ex-smoker & X & X & cM1 & 2.43 to 16.90 \\
        3 & Male & X & X & X & X & 9.07 to 34.50 \\
        \bottomrule
    \end{tabular}%
}
\end{table}

\begin{algorithm}
\caption{Pseudo-code for Human-in-the-Loop for Prognosis (HuLP) Model}
\begin{algorithmic}
\State \textbf{Given} a set of clinical features $P$; $m$ number of unique concepts in each clinical feature; input image $x$.
\State
\State Let $M = \sum_{j=1}^{|P|} m_j$. 
\State
\State \textbf{\underline{Encoder}}
    \State $y = \xi(x)$  \Comment{$x\in\mathbb{R}^{H\times W\times D\times C}, y\in\mathbb{R}^{(K\times M)}$}
    
\State
\For{$i$ in range($M$)}
    \State \textbf{\underline{Intervention}}
    \State $c_i = \alpha_i(y)$  \Comment{$c\in\mathbb{R}^{K}, K\%2=0$}
        \State $p_i=\sigma(c_i)$ \Comment{$p\in\mathbb{R}^{1}, p\in[0,1]$}
        \State During inference, and during training with a probability 
        \State of 0.25, $p_i$ is replaced with the ground-truth label \Comment{$p\in\mathbb{Z}^{1}, p\in[0,1]$}
    \State [$c_i^+, c_i^-] = c_i$ \Comment{$c^+, c^-\in\mathbb{R}^{K/2}$}
    \State $c_{Fi} = p_i c_i^+ + (1-p_i) c_i^-$ \Comment{$c_F\in\mathbb{R}^{K/2}$}

    \State
    \State \textbf{\underline{Classifier}}
    \State $q_i = \beta_i(c_{Fi})$  \Comment{$q\in\mathbb{R}^{1}$}
    \EndFor

\State
\State \textbf{\underline{Prognosticator}}
\State $w=[c_{F1}, c_{F2}, ..., c_{FM}]$
\State $y=\gamma(w)$ \Comment{$y\in\mathbb{R}^{n}$}

\end{algorithmic}
\label{pseudocode}
\end{algorithm}

\end{appendix}